\documentclass[pre,floats,aps,amsmath,amssymb, twocolumn]{revtex4}
\usepackage{graphicx}
\usepackage{url}
\usepackage{bm}
\usepackage{array}
\usepackage{multirow}
\usepackage[compact]{titlesec}

\begin{document}
\title{Exploring Correlation between Labels to improve Multi-Label Classification}
\author{Amit Garg, Jonathan Noyola, Romil Verma, Ashutosh Saxena, Aditya Jami \\ \textbf{Stanford University }}
\maketitle
\section{Abstract}
This paper attempts multi-label classification by extending the idea of independent binary classification models for each output label, and exploring how the inherent correlation between output labels can be used to improve predictions. Logistic Regression, Naive Bayes, Random Forest, and SVM models were constructed, with SVM giving the best results: an improvement of 12.9\% over binary models was achieved for hold out cross validation by augmenting with pairwise correlation probabilities of the labels. \\

\section{Introduction}

Multi-label classification is the set of classification problems where the output vector has a variable length. The average number of labels per review varies across datasets, and in general is a function of the semantics of the text rather than the syntax. The learning algorithm needs to estimate the number of labels and make the correct predictions. \\

Previously multi-label classification problems were solved using problem transformation techniques (converting the problem into binary classification problems per output label), or by adapting the algorithm to directly perform multi-label classification \cite{Tsoumakas}. This paper extracts correlation information between labels and factors the joint probabilities into the model. \\

\section{Task Definition}

Given large datasets of product reviews from Amazon and Twitter and manually labelled multi-label classifications for each (ground truth), the algorithm aims to predict all classifications for a review \cite{Jami}. We aim to make the algorithm portable across various datasets, i.e., training a model on amazon reviews and using it to classify tweets from Twitter.\\

\section{Dataset}

The Amazon dataset contains reviews for 400,000 products across 40,000 different categories. Among these, we focused on books and book subcategories. The Twitter dataset has 100,000 tweets. Both the datasets have the same schema, and each review and tweet has a unique ID, pre-pruned content, and a tree of all of the product labels for the review up to the Book category at the root. \\

\section{Theory}

Our approach augments the independent binary classification model. In addition to the individual probabilities, this also looks at the probability of labels occurring together. Hence the Inference algorithm models the formula -- \\
 $$
P\left( y^{(m)} | x^{(m)}\right) = \prod_{i}P\left( y_{i}^{(m)} | x^{(m)}\right) \prod_{j,k} P\left(y_{j}^{(m)}, y_{k}^{(m)}\right)^{\alpha}
$$ where \( y_{i} = \{0, 1\}, j,k  \in \{i | y_{i} = 1\},  0 \le \alpha \le 1  \) \\

\( P(y_{i}^{(m)} | x^{(m)}) \) represents the probability of the independent binary model for label \( i \) classifying as 0 or 1 given input \( x \). \(  \prod_{i}P\left( y_{i}^{(m)} | x^{(m)}\right) \) represents the assignment of 0 or 1 to all \( y_{i} \) that maximizes the joint probability.  \\

The pairwise correlation probability between each pair of labels is stored in the correlation matrix. This normalized rows of the matrix represent probability of a particular label's coupling with every other label except for itself. This probability is computed as a prior for the entire dataset and is not dependent on the input feature vector itself. It is represented by \( P(y_{j}^{(m)}, y_{k}^{(m)}) \).  \\

\( \prod_{j,k} P\left(y_{j}^{(m)}, y_{k}^{(m)}\right)^{\alpha} \) represents the joint probability of all pairwise combinations of predicted labels, each discounted by \( \alpha \). The probabilities obtained from the correlation matrix are given less weight than the probabilities obtained from independent models, hence the discount. This is because co-occurrence of labels  is not completely characterized by correlation, it also requires the higher order moments which significantly increase the computational overhead. \\

\section{Methodology}

The entire algorithm is split into 3 steps after parsing -- preprocessing, training independent classifiers, and incorporating co-occurrence probability to make the final label set. Hyper-parameters need to be tuned for each dataset. \\

LIBLINEAR works well for larger datasets while LIBSVM works well for custom and smaller datasets. LIBLINEAR only supports a linear Kernel but is very fast relative to LIBSVM and hence used in the learning algorithms \cite{Albanese}. \\

\subsection*{Parsing}

The Amazon and Twitter datasets are parsed to extract Book reviews and their corresponding labels (ground truth). The labels are structured in a hierarchy - for Books, there are 31 top level labels and for each label there are several sub-labels. For instance a top level label is "Literature and Fiction" and few sub labels associated with it are "Folklore", "Mysteries" and "Classics." The reviews are labelled with these sub-level labels which we map to one of the 31 top-level labels, and use for multi-label classification. \\

\subsection*{Preprocessing}

Training and testing of the algorithm is done on at most 3,000 reviews per dataset due to computational constraints during learning. Each Book review is mapped to a bag-of-words based input feature. On parsing a review, its content is pruned and stemmed and then a tf-idf based vector is generated which is used as the input feature \( x \). The labels corresponding to this review are associated as its outputs \( y \). \\

\begin{figure}[hb]
  \centering
  \includegraphics[width=3.2in]{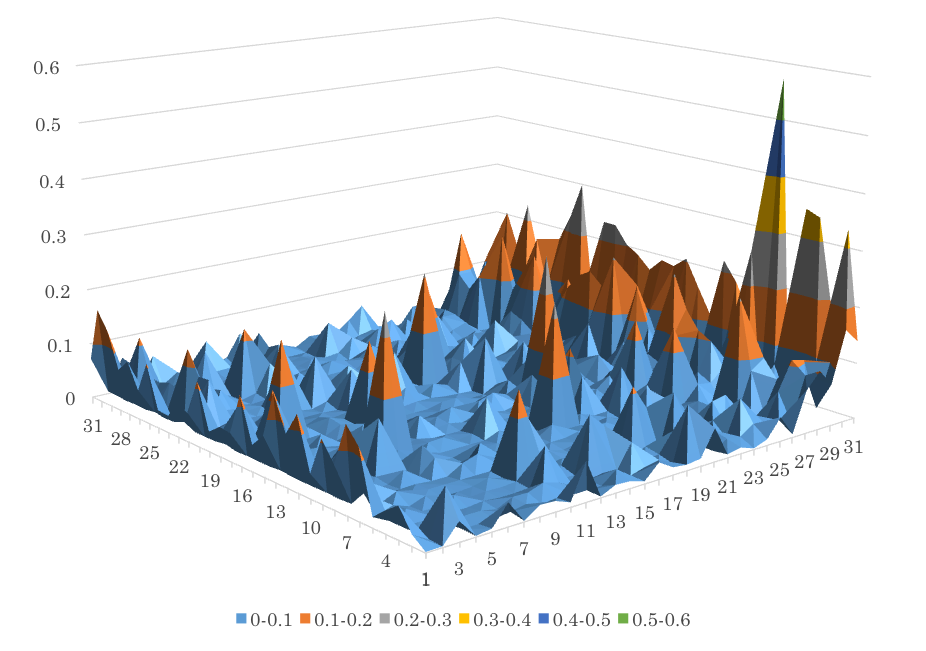}
  \caption[Correlation]
   {Correlation Matrix of 31 Book Labels from Amazon}
\end{figure}

The correlation matrix is built by parsing 100,000 Book reviews and creating pairwise counts of different labels -- a 31 \( \times \) 31 row normalized matrix with Laplace Smoothing. After normalization the matrix is not symmetric, so the geometric mean of cells \((i,j)\)and \((j,i)\) is treated as the actual correlation probability for two labels. Fig 1. plots the row normalized covariance matrix for Amazon dataset. The existence of large peaks is the motivation behind the algorithm.\\

\subsection*{Algorithm}

\begin{itemize}

\item The inputs to the learning algorithm are tf-idf based input features, their manually assigned labels, and the prior generated correlation matrix

\item 31 independent binary models are trained using L2-regularized SVM. This serves as the \textbf{baseline}.

\item For a review, the J labels with highest probabilities from the independent models are selected.

\item Probabilities of pairwise-combinations of these labels are computed as \( P\left(a\right)P\left(b\right) P\left(a, b\right)^{\alpha} \)

\item K \((a,b)\) pairs with maximum probability product are chosen and distinct labels constitute the predicted set.

\end{itemize}

The algorithm does not compute the theoretical model exactly but is an approximation. It is similar to beam search. \\

\subsection*{Hyperparameters}
\begin{table}[h]
    \begin{tabular}{| l | l | l | l |}
    \hline
    Hyper-parameter & For Amazon & For Twitter \\ \hline
    solver & L2-Regularized SVM   & L2-Regularized SVM   \\
    \( \alpha \)  & 0.3 & 0.25 \\
    \( \gamma \)  & 1/3 & 1/3 \\
    J & 12 & 10 \\
    K & 6 & count predicted by \\&& baseline \\
    \hline
    \end{tabular}
    \caption{Hyperparameters for Amazon Dataset}
\end{table}
~\\


\section{Error Metric}

The overall prediction error can be seen as a combination of two different kinds of errors - false positives and false negatives. False positives are the labels that are in the predicted label set but not in groundtruth, and false negatives are the labels present in groundtruth, but not in predicted label set.\\

We do not weigh the two identically, but rather give slightly higher importance to false negatives. The reason behind this is that over-predictions can be further controlled using added heuristics, but the labels missed can never be regained without looking at the entire output label set again.

\begin{eqnarray*}
 \text{Error} = 1 - \left(\frac{\text{Total labels correctly predicted}}{\text{Number of labels in ground truth}}\right)  \\
 \times \left(\frac{\text{Total labels correctly predicted}}{\text{Total labels predicted}}\right)^{\gamma}
\end{eqnarray*}

${\gamma}$ is a hyper-parameter which cannot be tuned since that would put it to ${\infty}$, and J and K as 31, i.e. output all the labels, irrespective of the input, since this gives zero error. \\

\section{Results \\}

\subsection*{Hold out cross validation with 70\%/30\% split \\}

\subsubsection{Train on Amazon and Test on Amazon Dataset}

Fig 2. gives the performance of our algorithm with the "31 independent models" as the dataset size is varied. As is evident, the algorithm extracts about 10\% improvement over the baseline, just by looking at the correlation matrix, after tuning the hyper-parameters. This thus confirms our assumption regarding the utility of the correlation matrix for multi-label classification with variable number of outputs. \\

\begin{figure}[!htbp]
	\centering
	\includegraphics[width=3.2in]{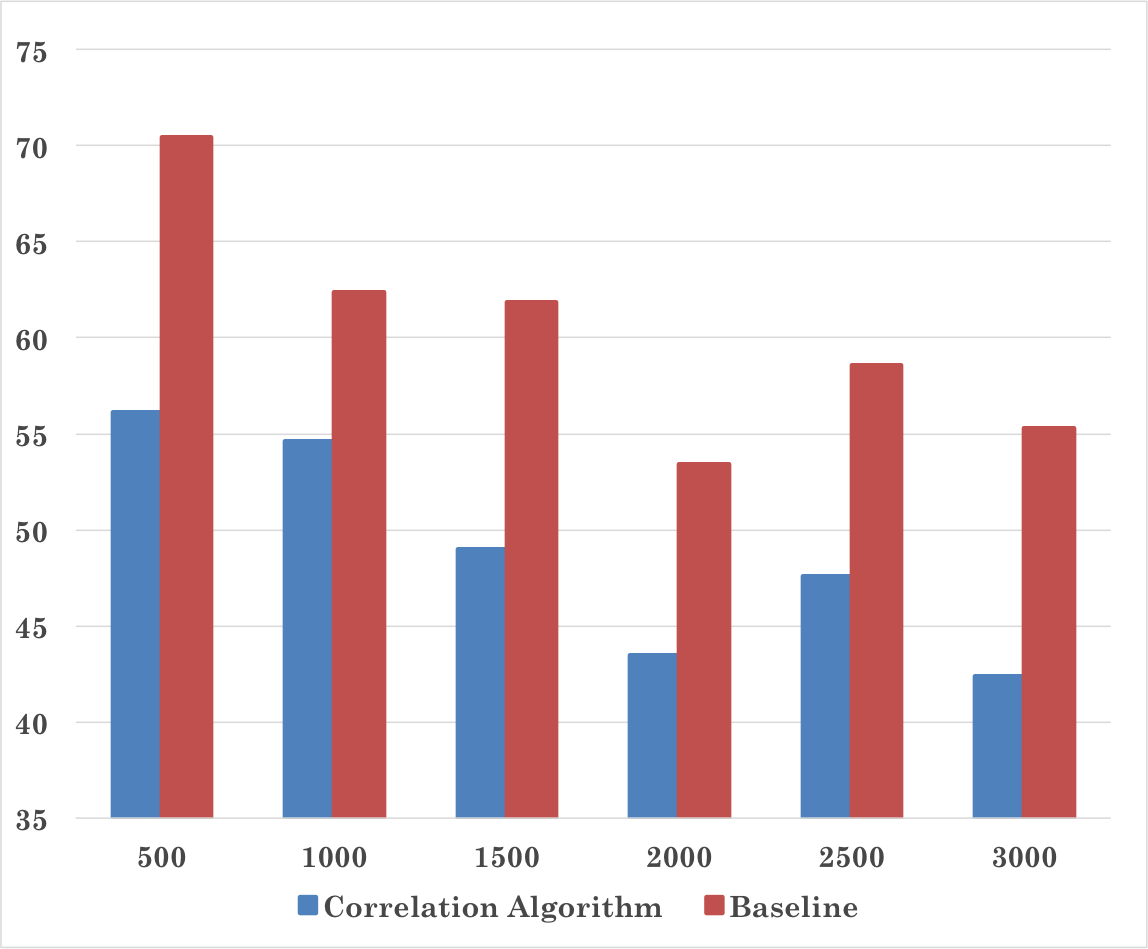}
	\caption[Correlation]
	{Percentage error with train, test on Amazon dataset}
\end{figure}

\subsubsection{Train and Test on Twitter Dataset}

Fig 3. is the performance of the algorithm when the Twitter dataset is used both for training and testing, after appropriately tweaking the hyper-parameters. The gain is only about 0.1\%, which can be attributed to the scarcity of content per Twitter review (the dataset size for Twitter was about 12\% of Amazon's for same number of reviews). Upon inspecting the output labels predicted by our algorithm, it was apparent that, the over predictions were the major contributors to the error, unlike Amazon where Misses were the prime error contributors. \\

\begin{figure}[!htbp]
	\centering
	\includegraphics[width=3.2in]{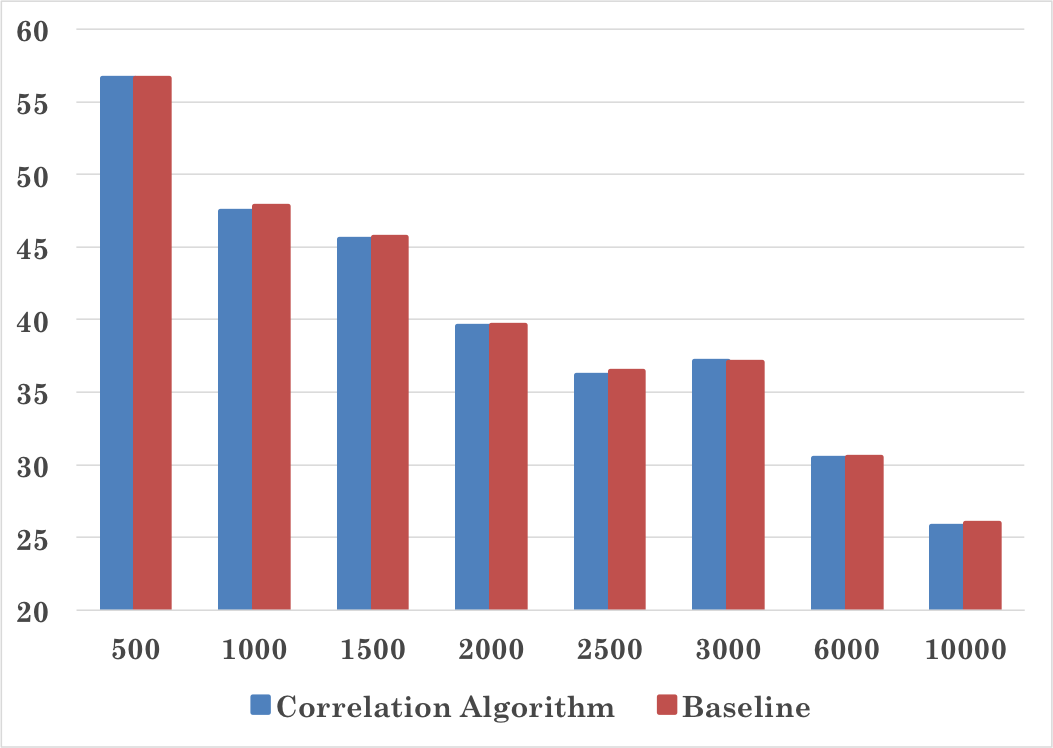}
	\caption[Correlation]
	{Percentage error while training and testing on Twitter dataset}
\end{figure}

\subsubsection{Train on Amazon and Test on Twitter Dataset}

For this case, we train on 100\% data from Amazon and test on 100\% data from Twitter. Fig 4. gives the performance of our algorithm when the training was done using Amazon's dataset and testing was done on Twitter's dataset. The hyper-parameters used are the same as those for Fig 2. Thus, the improvement observed is much smaller, the reason being the same, i.e., scarcity of content per tweet. \\

\begin{figure}[!htbp]
	\centering
	\includegraphics[width=3.2in]{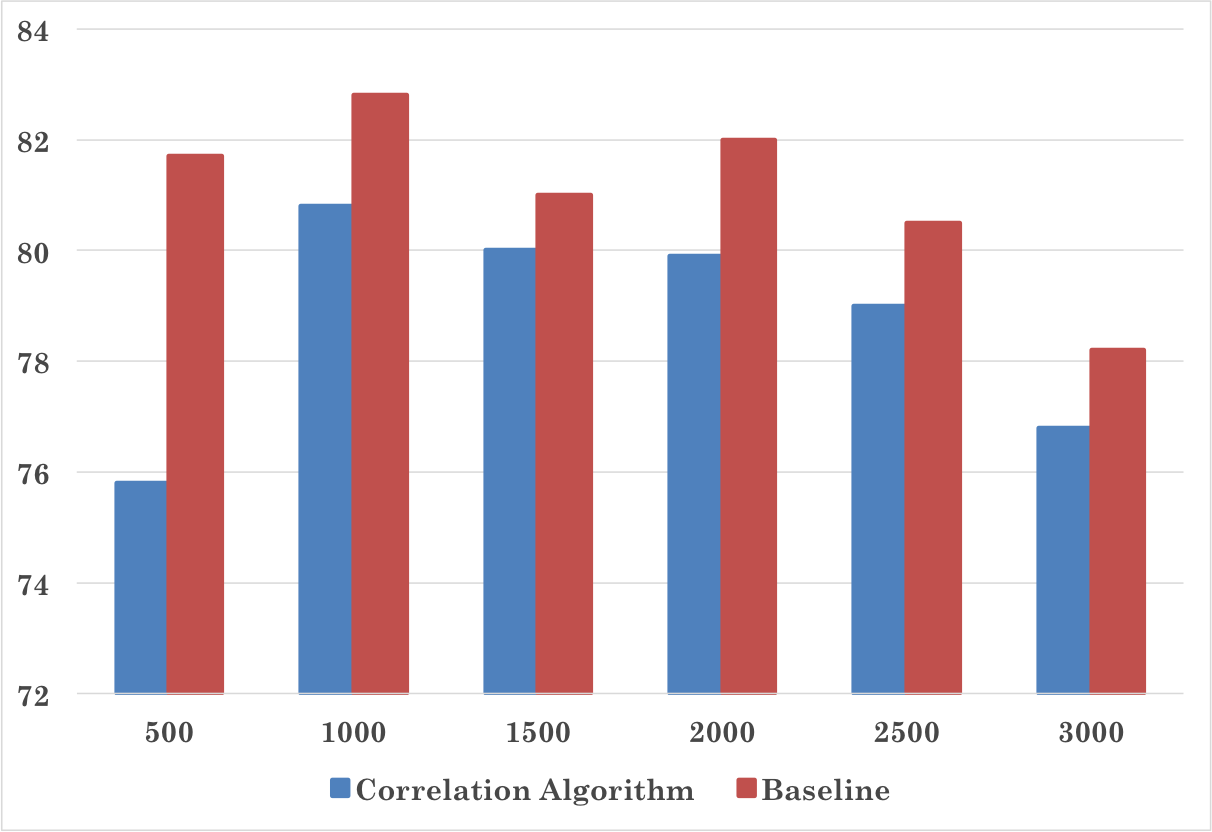}
	\caption[Correlation]
	{Percentage error while training on Amazon and testing on Twitter dataset}
\end{figure}

\subsection*{10\% K-fold Cross Validation}

K-fold (K=10) cross-validation revealed some important information regarding the algorithms' performance. While our algorithm had observed a 10 \% improvement over the collection of independent binary models for the Hold out cross validation, the same did not hold true for the results obtained from the K-fold cross validation. \\

Training and testing on Amazon no longer gave any improvement, but about a 2\% degradation in performance. The results for Twitter dataset also substantially changed, but since both the algorithms got similar shifts, the actual improvements were still marginal as before. \\

\begin{figure}[!htbp]
	\centering
	\includegraphics[width=3.2in]{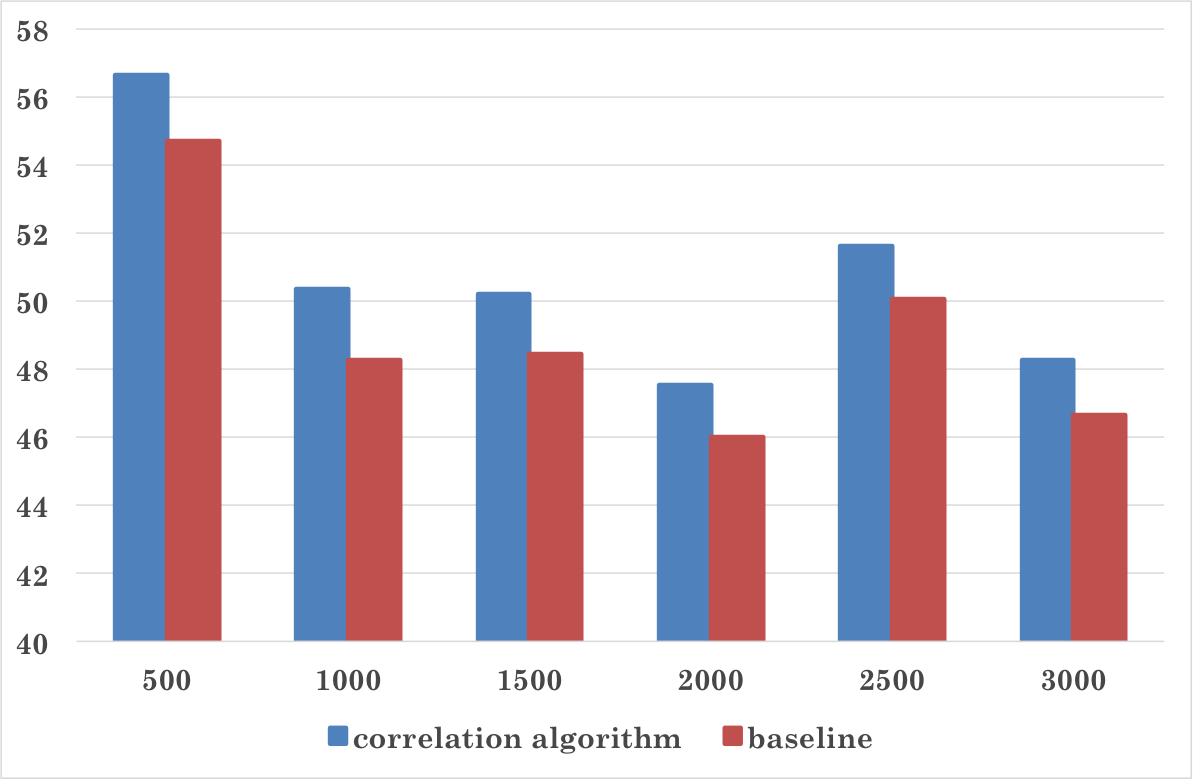}
	\caption[Correlation]
	{K-fold cross validation when training on Amazon and testing on Amazon dataset}
\end{figure}

We attribute the correlation algorthim's poor performance to two causes. First, Amazon dataset's number of labels per review has a mean of 3.06 and a standard deviation of 1.3. Because the standard deviation is reasonably high, it can be expected that the average number of labels per review for a given partition may be quite different from the global average. This causes our algorithm to produce many false positives when testing on some partitions where the number of labels per review is much lower than average. \\

Also, because our fixed correlation matrix is calculated from 100,000 samples tending towards the true correlation values, some test data partitions are likely to deviate from this average (highly biased test datasets), causing our algorithm to predict based on the true correlation, while the baseline algorithm ignores correlation and trains only on the training data. This hypothesis is supported by smoothing out the bias by increasing K, in which case the two algorithms' errors converge. \\

\begin{figure}[!htbp]
	\centering
	\includegraphics[width=3.2in]{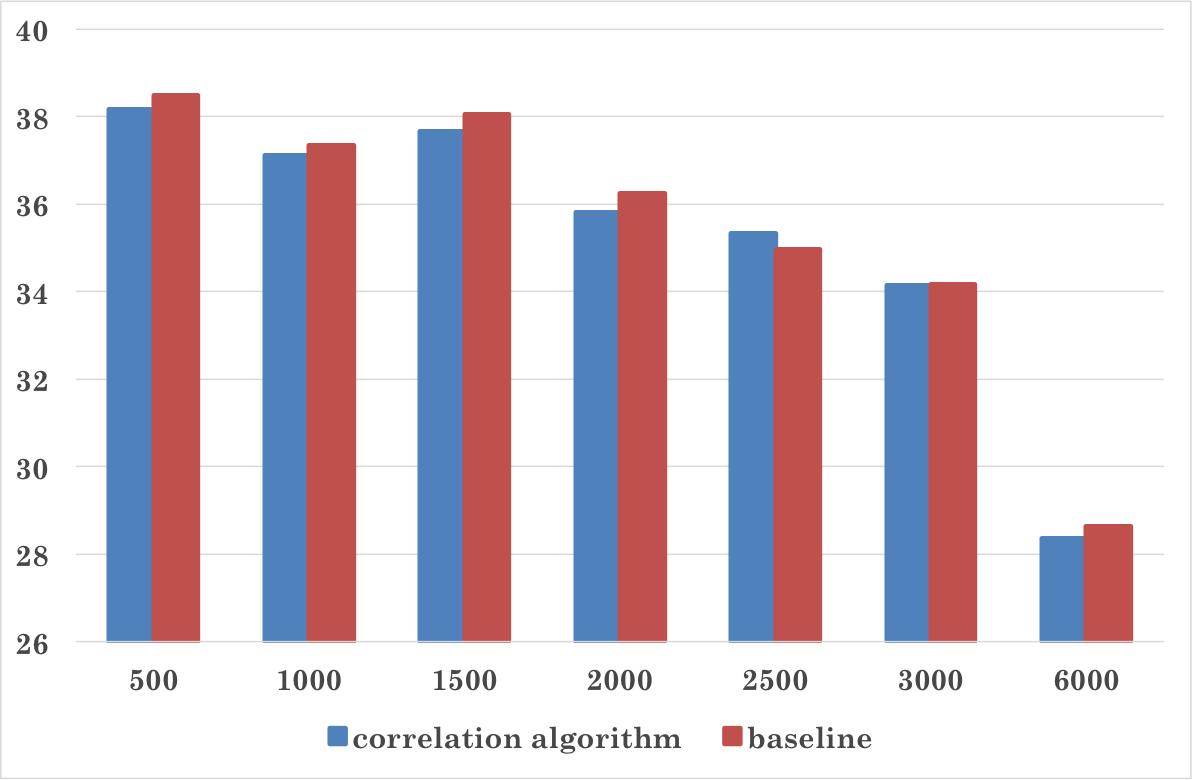}
	\caption[Correlation]
	{K-fold cross validation for train and test on Twitter}
\end{figure}

\subsection*{Comparing Learning Algorithms}

We compare the baseline results of three different models: SVM, Naive Bayes, and Random Forest. Naive Bayes did not scale well, but converged quickly. Additionally, Naive Bayes in Scikit-Learn \cite{Pedregosa} returned binary probabilities which do not lend themselves well to multi-label classification where each review has a variable number of labels. SVM however, had a noticeable trend and scaled best with increasing dataset. 

\begin{figure}[!htbp]
	\centering
	\includegraphics[width=3.2in]{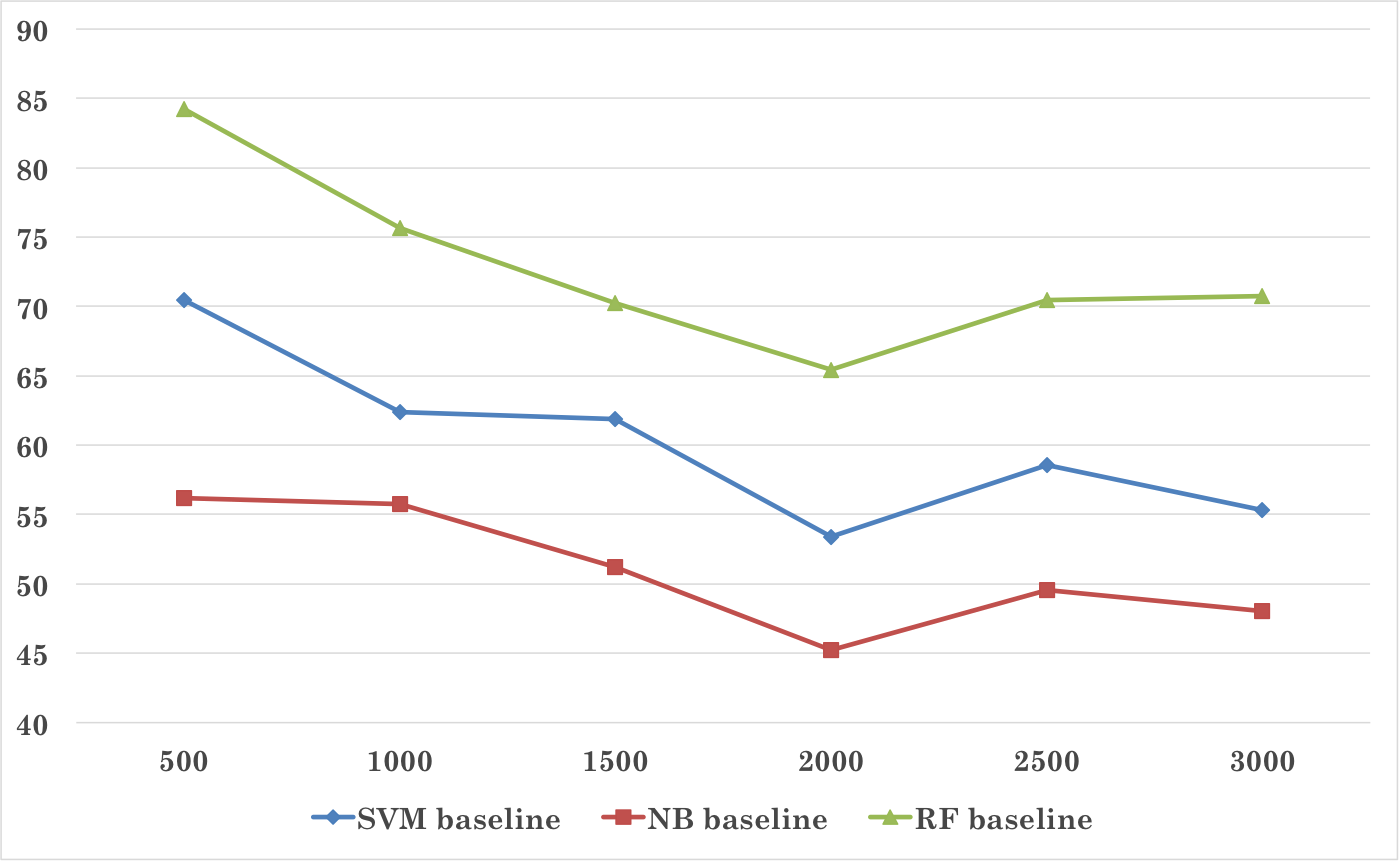}
	\caption[Correlation]
	{Baseline error for SVM, Naive Bayes, and Random Forest}
\end{figure}

Then each model was augmented with the correlation matrix. SVM gave best results and was then optimized by adjusting hyper-parameters. \\

\begin{figure}[!htbp]
	\centering
	\includegraphics[width=3.2in]{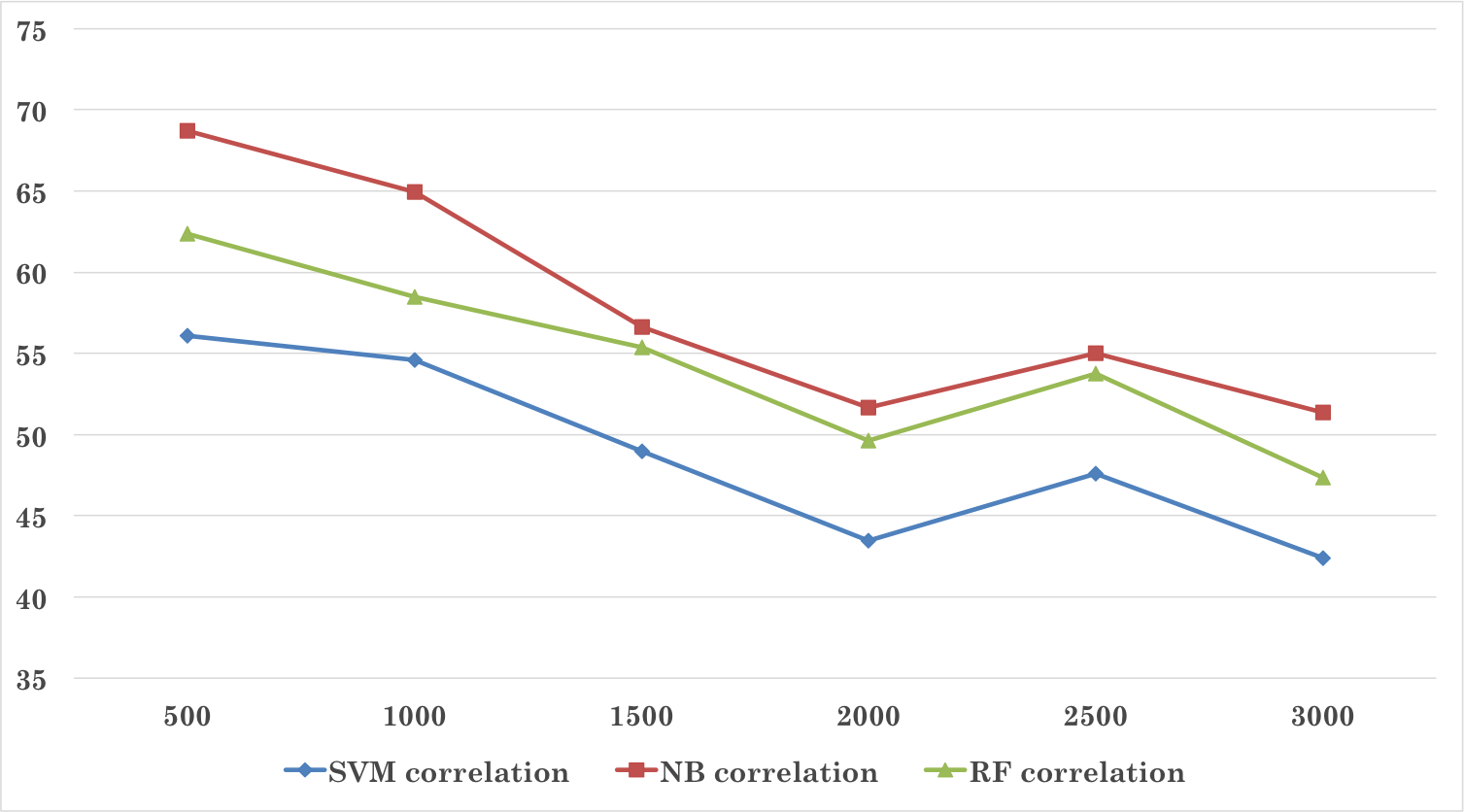}
	\caption[Correlation]
	{Correlation algorithm error for SVM, Naive Bayes, and Random Forest}
\end{figure}

\subsection*{Other Approaches Considered}
Another approach was to train \( ^{31}C_{2} = 465 \) (per label pair) separate correlation models. This conditioned the correlation probability on the train data instead of being fixed for a dataset. This increased error slightly and the computational time greatly. This method did not perform as well because, in the new correlation models the error to match an input vector to a label got compounded in with the correlation error. As a fix, the algorithm weighed both the correlation probability by $\alpha$ and independent probabilities by $1 - \alpha$. This marginally reduced the error hence did not continue further. \\

\section{Analysis \\}
\subsection*{Precision, Recall and F1 Score}
Train and test on 3000 Amazon reviews \( \beta = 1/3 \) 
\begin{table}[!h]
\begin{tabular}{ccccc}
                                             & \multicolumn{4}{c}{Predicted}                                                                                                                                                                                                                                                                       \\
                                             & \multicolumn{2}{c}{Correlated}                                                                                                                   & \multicolumn{2}{c}{Baseline}                                                                                                                     \\ \cline{2-5} 
\multicolumn{1}{c|}{\multirow{2}{*}{Actual}} & \multicolumn{1}{>{\centering\arraybackslash}p{0.7in}|}{\begin{tabular}[c]{@{}c@{}}TP 2041 \\\end{tabular}} & \multicolumn{1}{>{\centering\arraybackslash}p{0.7in}|}{\begin{tabular}[c]{@{}c@{}}FN 980\end{tabular}}   & \multicolumn{1}{>{\centering\arraybackslash}p{0.7in}|}{\begin{tabular}[c]{@{}c@{}}TP 1447\end{tabular}} & \multicolumn{1}{>{\centering\arraybackslash}p{0.7in}|}{\begin{tabular}[c]{@{}c@{}}FN 1574\end{tabular}}   \\ \cline{2-5} 
\multicolumn{1}{c|}{}                        & \multicolumn{1}{c|}{\begin{tabular}[c]{@{}c@{}}FP 2354 \\ \end{tabular}}  & \multicolumn{1}{c|}{\begin{tabular}[c]{@{}c@{}}TN 25315\end{tabular}} & \multicolumn{1}{c|}{\begin{tabular}[c]{@{}c@{}}FP 705\end{tabular}}  & \multicolumn{1}{c|}{\begin{tabular}[c]{@{}c@{}}TN 26964\end{tabular}} \\ \cline{2-5} 
\end{tabular}
\caption{Confusion Matrix for Train, Test on Amazon }
\end{table}

\begin{table}[!h]
\begin{tabular}{c|c|c|c|c|}
\cline{2-5}
                                  & \multicolumn{1}{>{\centering\arraybackslash}p{0.6in}|}{Precision} & \multicolumn{1}{>{\centering\arraybackslash}p{0.6in}|}{Recall} & \multicolumn{1}{>{\centering\arraybackslash}p{0.6in}|}{\(F_{1}\)}     & \multicolumn{1}{>{\centering\arraybackslash}p{0.6in}|}{\(F_{\beta}\)} \\ \hline
\multicolumn{1}{|c|}{Baseline}    & 0.6724    & 0.4789 & 0.5594 & 0.4932 \\ \hline
\multicolumn{1}{|c|}{Correlation} & 0.4644    & 0.6756 & 0.5504 & 0.6421 \\ \hline
\end{tabular}
\caption{Scores for Train, Test on Amazon Dataset }
\end{table}

Train and test on 3000 Twitter reviews, \( \beta = 1/3 \)

\begin{table}[!h]
\begin{tabular}{ccccc}
                                             & \multicolumn{4}{c}{Predicted}                                                                                                                                                                                                                                                                       \\
                                             & \multicolumn{2}{c}{Correlated}                                                                                                                   & \multicolumn{2}{c}{Baseline}                                                                                                                     \\ \cline{2-5} 
\multicolumn{1}{c|}{\multirow{2}{*}{Actual}} & \multicolumn{1}{>{\centering\arraybackslash}p{0.7in}|}{\begin{tabular}[c]{@{}c@{}}TP 1124 \\ \end{tabular}} & \multicolumn{1}{>{\centering\arraybackslash}p{0.7in}|}{\begin{tabular}[c]{@{}c@{}}FN 658\end{tabular}}   & \multicolumn{1}{>{\centering\arraybackslash}p{0.7in}|}{\begin{tabular}[c]{@{}c@{}}TP 1124\end{tabular}} & \multicolumn{1}{>{\centering\arraybackslash}p{0.7in}|}{\begin{tabular}[c]{@{}c@{}}FN 658\end{tabular}}   \\ \cline{2-5} 
\multicolumn{1}{c|}{}                        & \multicolumn{1}{c|}{\begin{tabular}[c]{@{}c@{}}FP 433 \\ \end{tabular}}  & \multicolumn{1}{c|}{\begin{tabular}[c]{@{}c@{}}TN 25716\end{tabular}} & \multicolumn{1}{c|}{\begin{tabular}[c]{@{}c@{}}FP 382\end{tabular}}  & \multicolumn{1}{c|}{\begin{tabular}[c]{@{}c@{}}TN 25767\end{tabular}} \\ \cline{2-5} 
\end{tabular}
\caption{Confusion Matrix for Train, Test on Twitter }
\end{table}

\begin{table}[!h]
\begin{tabular}{c|c|c|c|c|}
\cline{2-5}
                                  & \multicolumn{1}{>{\centering\arraybackslash}p{0.6in}|}{Precision} & \multicolumn{1}{>{\centering\arraybackslash}p{0.6in}|}{Recall} & \multicolumn{1}{>{\centering\arraybackslash}p{0.6in}|}{\(F_{1}\)}     & \multicolumn{1}{>{\centering\arraybackslash}p{0.6in}|}{\(F_{\beta}\)} \\ \hline
\multicolumn{1}{|c|}{Baseline}    & 0.7464    & 0.6308 & 0.6837 & 0.6407 \\ \hline
\multicolumn{1}{|c|}{Correlation} & 0.7219    & 0.6306 & 0.6733 & 0.6388 \\ \hline
\end{tabular}
\caption{Scores for Train, Test on Twitter Dataset }
\end{table}

Train and test on 3000 Amazon, Twitter reviews

\begin{table}[!h]
\begin{tabular}{ccccc}
                                             & \multicolumn{4}{c}{Predicted}                                                                                                                                                                                                                                                                       \\
                                             & \multicolumn{2}{c}{Correlated}                                                                                                                   & \multicolumn{2}{c}{Baseline}                                                                                                                     \\ \cline{2-5} 
\multicolumn{1}{c|}{\multirow{2}{*}{Actual}} & \multicolumn{1}{>{\centering\arraybackslash}p{0.7in}|}{\begin{tabular}[c]{@{}c@{}}TP 1976\\ \end{tabular}} & \multicolumn{1}{>{\centering\arraybackslash}p{0.7in}|}{\begin{tabular}[c]{@{}c@{}}FN 4013\end{tabular}}   & \multicolumn{1}{>{\centering\arraybackslash}p{0.7in}|}{\begin{tabular}[c]{@{}c@{}}TP 2223\end{tabular}} & \multicolumn{1}{>{\centering\arraybackslash}p{0.7in}|}{\begin{tabular}[c]{@{}c@{}}FN 3766\end{tabular}}   \\ \cline{2-5} 
\multicolumn{1}{c|}{}                        & \multicolumn{1}{c|}{\begin{tabular}[c]{@{}c@{}}FP 12833 \\ \end{tabular}}  & \multicolumn{1}{c|}{\begin{tabular}[c]{@{}c@{}}TN 74209\end{tabular}} & \multicolumn{1}{c|}{\begin{tabular}[c]{@{}c@{}}FP 23353\end{tabular}}  & \multicolumn{1}{c|}{\begin{tabular}[c]{@{}c@{}}TN 63689\end{tabular}} \\ \cline{2-5} 
\end{tabular}
\caption{Confusion Matrix for Train Amazon, Test Twitter }
\end{table}

\begin{table}[!h]
\begin{tabular}{c|c|c|c|c|}
\cline{2-5}
                                  & \multicolumn{1}{>{\centering\arraybackslash}p{0.6in}|}{Precision} & \multicolumn{1}{>{\centering\arraybackslash}p{0.6in}|}{Recall} & \multicolumn{1}{>{\centering\arraybackslash}p{0.6in}|}{\(F_{1}\)}     & \multicolumn{1}{>{\centering\arraybackslash}p{0.6in}|}{\(F_{\beta}\)} \\ \hline
\multicolumn{1}{|c|}{Baseline}    & 0.0869    & 0.3712 & 0.1409 & 0.2797 \\ \hline
\multicolumn{1}{|c|}{Correlation} & 0.1334    & 0.3299 & 0.1900 & 0.2876 \\ \hline
\end{tabular}
\caption{Scores for Training Amazon, Test Twitter }
\end{table}

The correlation algorithm weighs false positives and false negatives differently, hence the \( F_{\beta} \) score gives a more accurate understanding of the algorithm's performance. \( F_{\beta} \) for the correlation algorithm's test on Amazon is 30.19\% better than the baseline's test, 2.82\% better than the baseline's test on cross-domain learning, and 0.3\% worse than baseline's test on Twitter. This validates the trend in the plots. \\

The false positive counts when training and testing purely on one dataset are usually higher than the false positive counts from the baseline, as in our approach we tend to discount the error of making false positives given we meet true positives. \\

Table V shows the confusion matrix for cross domain learning where the correlation algorithm's true negative count is much higher than that of the baseline. By using the correlation between output labels, the algorithm discards labels that independently had a high probability but were not well correlated with other high probability labels. \\

\subsection*{Bias and Variance}

As evident, the observed training error for each dataset was much less than the observed test error, indicating high variance and low bias. To further investigate, Principal Component Analysis was used. Since the feature vectors were based on word occurrences in text, they were significantly larger than the dataset size -- the observed feature size for 2500 dataset size of Amazon reviews was 12000. Due to computational limitations, straightforward PCA was infeasible. \\

Sparse notation was used to represent feature vector, from Python's numpy library. Then Sparse SVD (singular value decomposition) found the smallest subspace that the Feature matrix mapped to, keeping all singular values greater than 1. This significantly reduced the dataset size. The 2500 dataset of Amazon reviews, now reduced to a feature vector of size 2300. While the number of features was still comparable to the number of data points, it was substantially lower than the size of the raw feature vector (1/6). \\

This experimentation exposed two facts. The errors were still the same, but the computation time rose significantly (about five times the previous duration). It was seen that LibLinear, the python library, was tuned to work with large datasets with document based feature vectors (as compared to LibSvm), and not for any general features. \\

Since the baseline itself suffers from the issue of high variance and low bias, any augmentations over it would be unable to resolve this by themselves, and would require modifying how the TF-IDF vectors are generated. We had already stemmed the words to check this issue, but evidently this would require a more careful processing of the reviews, before they are converted to features. Moreover, since SVMs enforce larger margin as a metric to evaluate each point, they end with a much lower VC dimension than the size of the feature vector. This was another reason in favor of SVM over Logistic Regression, Naive Bayes, and Random Forest.\\

The baseline resulted in about 10\% training error, while our algorithm resulted in about 15\% training error. Thus there is an increased bias, although the higher variance is a more pressing issue because the test error is significantly higher than training error for both algorithms. \\

\section{Conclusion}

Using the correlation matrix to help classify reviews appeared to be a good approach, justified by peaks in correlation. The results show, however, that when performing multi-label classification with a sufficiently sized feature set, augmentation of an independent model with second order correlation probability shows only marginal improvements.\\

The model was modified from the baseline by adding a single feature: second order correlation. This is the greedy method of selecting features, in contrast to exploring correlation of all combinations of labels. In such a case where we did not explore higher order correlation we could have missed, for example, that labels A, B, and C never occur together -- a piece of information that could have proved vital to the model's representation of the true data. Thus, if computation time and space are not an issue, higher order correlation should be considered. \\

Additionally, SVM, as was already known, has shown to be best for learning on large feature sets because it will attempt to reduce the set and hence generalize better. \\

\section{Challenges \& Future Work}

The correlation matrix was built specific for every dataset, so to apply the learning algorithms on cross datasets a generalized correlation framework is needed. \\

Currently the hyper-parameter K is optimized to choose number of predicted labels using supervised techniques. The number of labels for each review could be predicted by using unsupervised techniques such as k-means clustering on label probabilities. \\

The datasets have different inherent structures which do not generalize well when using a common learning algorithm. For the Twitter dataset, each feature vector is about 12\% the size of a corresponding feature vector in the Amazon dataset. The features should be supplemented with options such as semantic relations in text to make classifications and context of user history. \\

The reviews have sub-level labels associated with them which are rolled up to first-level labels to make the classifications. The algorithm should extend to make hierarchical classifications.\\

\section{Acknowledgments}

Sincere thanks to Professor Ashutosh Saxena and to Aditya Jami for advising the research team and providing the labelled datasets of product reviews from Amazon and Twitter.


\end{document}